\documentclass[a4paper,11pt,twocolumn,twoside]{article}
\usepackage[]{graphicx}
\usepackage{sepln_en}
\usepackage{fullname}
\usepackage[utf8]{inputenc}
\usepackage[english]{babel}
\usepackage{booktabs}
\usepackage{xcolor}
\usepackage[most]{tcolorbox}
\usepackage{multirow}
\usepackage{csquotes}

\input epsf

\newcommand{\system}{System}
\newcommand{\model}{Model}

\newcommand{\blind}{Blind}
\newcommand{\allcontext}{All-in-context}
\newcommand{\ICLSieben}{ReAct ICL 70B}
\newcommand{\FTSieben}{ReAct FT 70B}
\newcommand{\ICLAcht}{ReAct ICL 8B}
\newcommand{\FTAcht}{ReAct FT 8B}
\newcommand{\UST}{USI ReAct 8B}
\newcommand{\External}{BART}

\title{Learning to Reason and Use Tools through Unsupervised Fine-Tuning in Task-Oriented Dialog Systems}

\author {\textbf{Markel Ferro,$^1$} \textbf{Oier Lopez de Lacalle$^1$}\\
$^1$HiTZ Center - Ixa, University of the Basque Country UPV/EHU\\
\{markel.ferro, oier.lopezdelacalle\}@ehu.eus\\
}

\seplntranstitle{Aprendiendo a razonar y usar herramientas mediante fine-tuning no supervisado en sistemas de diálogo orientados a tareas}

\seplnclave{Sistemas de Diálogo, Generación de Lenguaje Natural, Aprendizaje No Supervisado.}

\seplnresumen{Los sistemas de diálogo actuales tienen problemas con la recuperación dinámica de información, causando alucinaciones y menor precisión. Abordamos este problema adaptando ReAct al Diálogo Orientado a Tareas, permitiendo a los Grandes Modelos de Lenguaje (LLM) usar conocimiento externo y generar respuestas fácticas. Proponemos un proceso de fine-tuning no supervisado que recopila trayectorias de razonamiento mediante inferencia de modelos adaptados en contexto. Un juez LLM filtra las muestras de alta calidad para construir un conjunto de entrenamiento robusto. Esto se potencia con un bucle de automejora no supervisado, donde los checkpoints refinados generan trayectorias superiores para iteraciones posteriores. Experimentos en SIMMC demuestran que los sistemas ReAct superan a los modelos de referencia gracias a un mejor razonamiento y uso de herramientas. Destacamos que nuestro modelo de 8B ajustado supera a un sistema en contexto de 70B. Finalmente, analizamos los errores, la complejidad de la escena y la generalización multidominio.}

\seplnkey{Dialog Systems, Natural Language Generation, Unsupervised Learning.}

\seplnabstract{Current dialogue systems struggle with dynamic information retrieval, often leading to hallucinations and lower response accuracy. We address this by adapting the ReAct framework for Task-Oriented Dialogue, enabling Large Language Models (LLMs) to access external knowledge and produce factual responses. Mainly, we propose an unsupervised fine-tuning pipeline that harvests reasoning trajectories via in-context learning inference. High-quality samples are filtered using an LLM-based judge to construct a robust training set. This is enhanced by a unsupervised self-improvement loop, where improved checkpoints generate increasingly better trajectories for subsequent fine-tuning iterations. Experiments on the SIMMC dataset demonstrate that ReAct-based systems outperform baselines due to superior reasoning and tool use. Notably, our fine-tuned 8B model surpasses a 70B in-context system. Finally, we present an error analysis, impact of scene complexity, and cross-domain generalization.}

\firstpageno{1}

\begin{document}

\label{firstpage} \maketitle

\section{Introduction}

Task-oriented Dialogue (TOD) systems aim to accomplish predefined tasks, such as booking a flight or retrieving product information, through natural language interactions with users ~\cite{wang2025surveyevolutionlanguagemodelbased}. Traditionally, these systems followed a modular architecture consisting of separate components for Natural Language Understanding (NLU), Dialogue State Tracking (DST), and Natural Language Generation (NLG), among others~\cite{Celikyilmaz2018}.
While effective, modular approaches are inherently complex, susceptible to error propagation between components, and require independent training of each module. 
Consequently, recent advances in Large Language Models (LLMs) have driven a shift toward end-to-end TOD systems, where generic LLMs are adapted to construct task-specific conversational agents~\cite{yi2025surveyrecentadvancesllmbased}. 
Although such models produce fluent and natural responses, they frequently exhibit factual inconsistencies, commonly referred to as hallucinations ~\cite{dhuliawala2023chainofverificationreduceshallucinationlarge}, 
as they rely solely on the knowledge obtained during training, and lack mechanisms for dynamically accessing real-time information necessary for accurate task completion.

To overcome these limitations, recent research has demonstrated that integrating Chain-of-Thought (CoT) reasoning~\cite{wei2023chainofthoughtpromptingelicitsreasoning} with tool-use in LLMs leads to strong performance in tasks requiring access to external information~\cite{yao2022react}. In particular, the ReAct framework employs carefully designed prompting strategies to guide the model through an iterative sequence of \textit{thoughts}, \textit{actions}, and \textit{observations}. Here, thoughts represent the model’s internal reasoning process, decomposing a task into subproblems; actions denote external API calls or program executions aimed at solving those subproblems; and observations capture the responses returned by the external tools.

However, adapting the ReAct framework to TOD systems is non-trivial~\cite{elizabeth-etal-2025-exploring}, as it requires the design of appropriate tools (i.e., actions) and carefully crafted prompting strategies that enable the model to reason effectively, decomposing tasks into subproblems and selecting the correct tools for each. In this paper, we show that such an adaptation is not only feasible but can be improved through a fully unsupervised fine-tuning methodology that enhances the agent’s reasoning and tool-use capabilities without human supervision. Our proposed pipeline first derives reasoning trajectories from the training data via In-Context Learning (ICL) inference. The generated trajectories are then filtered using an LLM-based judge, which selects only high-quality instances to construct a dataset for fine-tuning. This approach addresses the starting problem that agents have by autonomously generating and curating its own dataset.

Our evaluation is conducted manually on the SIMMC dataset~\cite{moon2020originalSIMMC,kottur-etal-2021-simmc,kottur-moon-2023-overview}, which comprises dialogues centered around objects situated in store environments. We analyze its fashion and furniture domains separately to assess how scene complexity affects system performance. To further investigate how difficulty affects performance, we introduce synthetic, \textit{fake}, objects to simulate scenarios with increased complexity. In addition, we examine how inefficient reasoning trajectories impact the fine-tuning process in terms of both performance and computational efficiency. Finally, we evaluate the generalization ability of our \MakeLowercase{\system} by comparing it against alternative approaches on the same dataset tasks. As a result of our experimentation, we find that:

\paragraph{1) Reasoning and tool use enhance information access in TOD systems.} We demonstrate that incorporating explicit reasoning and tool-use mechanisms enables more effective utilization of external knowledge resources compared to approaches where information access is limited to the input context.
\paragraph{2) Unsupervised iterative self-improvement effectively adapts smaller models.}Our proposed fine-tuning pipeline enables 8B-parameter models to match or even surpass the performance of their 70B-parameter counterparts. This result is particularly significant, as it provides the deployment advantages of smaller models (reduced computational cost and resource requirements) without compromising performance.
\paragraph{3) The proposed approach exhibits robust performance in complex scenarios.} In settings with high scene complexity, characterized by a large number of objects, prompting-based methods experience a more rapid degradation in performance compared to our ReAct-based approach.
\paragraph{4) Fine-tuning on reasoning trajectories yields strong cross-domain generalization.} We demonstrate that previous fine-tuning approaches often achieve in-domain performance through overfitting, resulting in poor transfer to unseen domains. In contrast, our methodology promotes better generalization across domains, effectively mitigating overfitting.

\begin{figure*}[!ht]
\begin{center}
    \includegraphics[width=\textwidth]{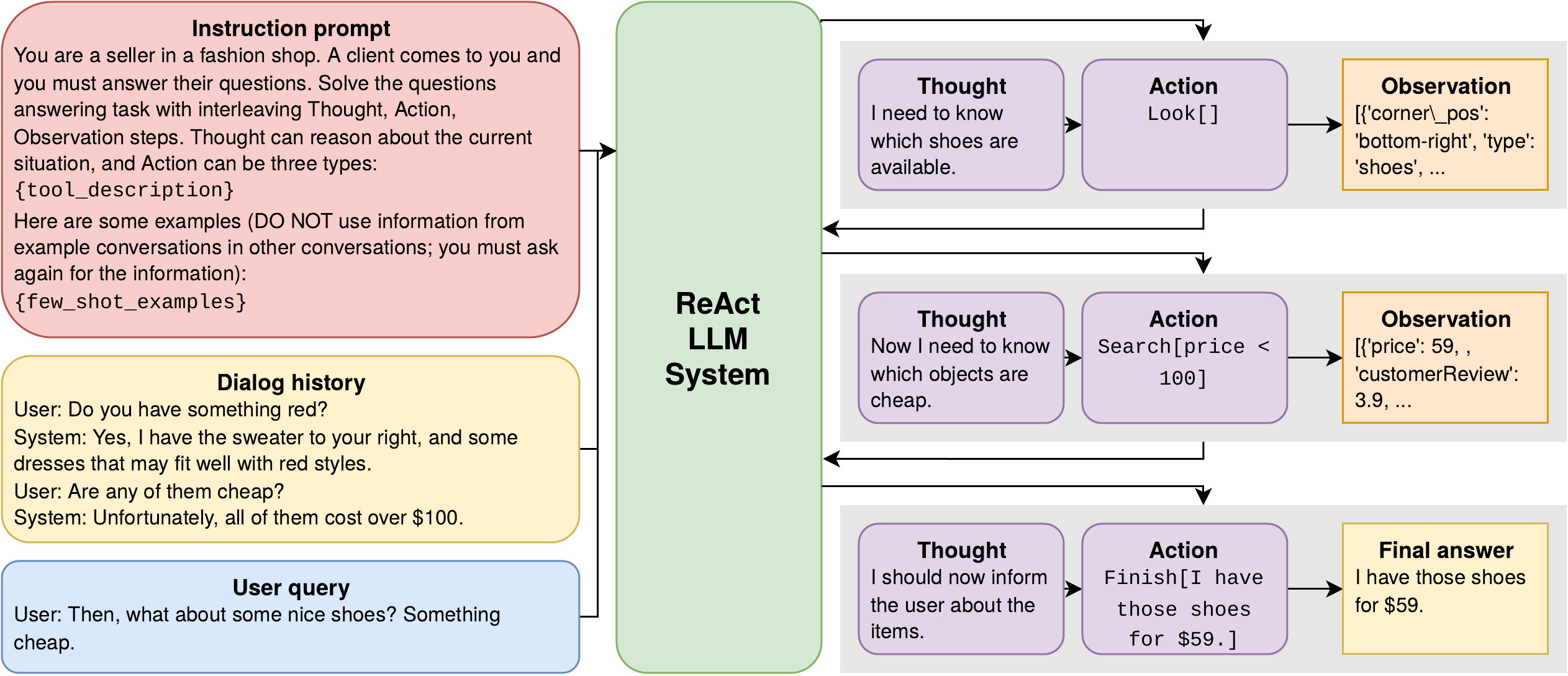}
    \caption{A diagram illustrating the ReAct LLM system as it processes a task. The system takes an Instruction prompt, Dialog history, and a User query as inputs. It then generates the Final Answer through an iterative sequence of Thoughts, Actions, and Observations.}
    \label{fig:general-diagram}
\end{center}
\end{figure*}

\section{Related Work}

\paragraph{TOD systems with agents} The rapid advancement of Large Language Models has been transformative. However, LLMs operate on static, parametric knowledge acquired during training, which results in them generating factually inconsistent or outdated information, a phenomenon known as ``hallucination''. This is particularly problematic in TOD systems, where providing correct information is necessary for task completion.
To mitigate these shortcomings, researchers have developed agentic frameworks, such as the ReAct framework~\cite{yao2022react}, that provide LLMs access to external information sources.
Recent research has focused on developing agentic TOD systems capable of reasoning, planning, and using tools~\cite{xu-etal-2024-rethinking}, with some works expanding on this idea by making use of multi-agent architectures~\cite{gupta2024dardmultiagentapproachtaskoriented} for managing the complexity of real-world dialogue.
However, despite the clear potential of agentic frameworks like ReAct, their direct application to the structured and constrained nature of TOD is not trivial~\cite{elizabeth-etal-2025-exploring}, as they have a tendency for the model to simply imitate the few-shot examples in its prompt, and to generate inconsistent reasoning. 

\paragraph{Evaluation of TOD systems} The evaluation of LLM dialogue systems has long been a notoriously difficult problem. Many traditional automatic metrics were inherited from other Natural Language Processing tasks like machine translation. Metrics such as BLEU~\cite{papineni-etal-2002-bleu} and ROUGE~\cite{lin-2004-rouge}, which measure the overlap between a generated response and a reference, are inadequate for assessing the quality of a conversation~\cite{liu-etal-2016-evaluate}. Even metrics designed for TOD, such as Inform Rate and Success Rate, have demonstrated significant limitations with LLMs. These metrics focus on the final outcome, e.g., whether the system provided the correct entity. The evaluations mainly rely on string-matching, which can lead to incorrect evaluations if the model contains the correct attribute somewhere else in the response, even if syntactically it does not refer to the correct object~\cite{acikgoz-etal-2025-td}. This is further exacerbated in multi-object responses. A methodology guided using these metrics may lead to systems that appear successful on paper but unreliable in real-world scenarios.

\paragraph{LLM judges} To address the problems of traditional metrics while avoiding the costly human evaluation, the research community has recently started using LLMs as judges. This approach aims to use the language understanding of LLMs to assess the quality of outputs generated by other models. Large proprietary models like GPT-4 are often employed for this purpose due to their high alignment with human evaluation~\cite{gu2025surveyllmasajudge}; however, their high cost and lack of controllability have spurred the development of open-source LLM judges.
A state-of-the-art example is Prometheus 2~\cite{kim2024prometheus2opensource}, an open-source model developed specifically for the task of evaluating other LLMs. It offers a high degree of controllability, supporting evaluation based on custom, user-defined criteria. It is capable of performing both direct assessment (assigning a score on a Likert scale) and pairwise ranking (determining a preference between two outputs).

\paragraph{Task-oriented Dialogue Datasets} SIMMC~\cite{kottur-etal-2021-simmc} and MultiWOZ~\cite{eric2019multiwoz21consolidatedmultidomain} are prominent reference datasets in the field. They provide multi-turn conversations focused on particular tasks, ranging from simulating a police interaction to purchasing clothes. However, despite their varied nature, models trained on these datasets expose the need for dynamic reasoning capabilities~\cite{hemanthage2023learnmetadata}, as they are often unable to adapt to new tasks. This limitation has motivated recent interest in strategies that dynamically adapt models to novel situations.

\section{ReAct as Dialogue System} 
\label{sec:react-dialogue-system}

\paragraph{Problem formulation} We base our study on the SIMMC 2.1 dataset~\cite{kottur-moon-2023-overview}, a task-oriented dialogue corpus designed for multimodal assistant–user interactions in realistic shopping environments. Although the benchmark defines four tasks, including coreference resolution and dialogue state tracking, this work focuses exclusively on the Automatic Response Generation task. The assistant does not have access to any ground-truth annotations, observing only the dialogue history, the current user utterance (current query) and the metadata of objects present in the scene.

The task is designed to evaluate a model’s ability to generate contextually appropriate responses to user queries that require information about relevant objects within a scene. Each dialogue occurs in a specific scene containing multiple objects, and the model has access to the corresponding object metadata. This metadata includes both visual and textual representations, and the \MakeLowercase{\system} must accurately perform metadata grounding by identifying and leveraging the objects relevant to the dialogue context.

\begin{figure*}[!th]
\begin{center}
    \includegraphics[width=\textwidth]{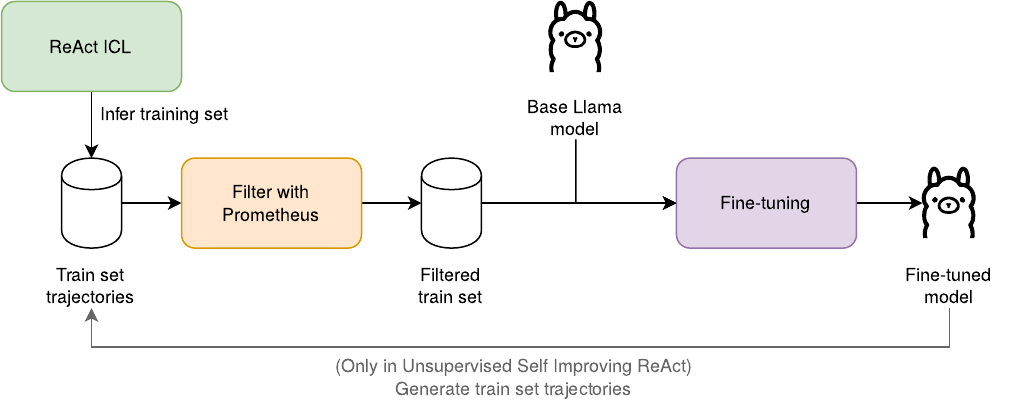}
    \caption{Diagram of the Fine-Tuning pipeline, which operates without human intervention by using model-generated instances for Unsupervised Self Improvement.}
    \label{fig:ft-diagram}
\end{center}
\end{figure*}

\paragraph{ReAct} %
To address this task, we design an agentic system in which the agent actively retrieves information about relevant objects. This design grounds the agent’s behavior, preventing unnecessary access to unrelated information. We implement this through an adaptation of the ReAct framework~\cite{yao2022react} tailored to our dataset and equip the agent with a set of custom tools specialized for information retrieval. Figure~\ref{fig:general-diagram} illustrates the adaptation of the ReAct framework to a dialogue system, where the LLM (depicted as the ReAct LLM system in the figure) receives the instruction prompt, dialogue history, and user query as inputs, and generates the final response through an iterative sequence of thoughts, actions, and observations triplets.
Actions require the use of external tools. The tools defined in our adaptation are as follows:

\begin{itemize}
    \item \texttt{\textbf{Look[]}}: %
    It returns the visual metadata of all objects present in the scene in textual form, simulating the capabilities that a multimodal model would possess.
    \item  \texttt{\textbf{Search[query]}}: %
    A tool that retrieves the non-visual metadata of objects in the scene that satisfy the constraints specified in the query. For example, \texttt{Search[customerRating>3]} would retrieve objects with ratings greater than 3. 
    \item \texttt{\textbf{Finish[answer]}}: %
    An action used by the model to indicate the final response that the system should return. Once this action is executed, no further actions are performed. For example, \texttt{Finish["That jackets costs \$300."]}.
\end{itemize}

Figure~\ref{fig:general-diagram} illustrates an example in which the user asks for inexpensive shoes. The \textit{thought-action-observation} triplets generated by the models (trajectory) are as follows: The \MakeLowercase{\system} first visualizes the available shoes using the \texttt{Look[]} tool, then, it searches for cheap items in the store using \texttt{Search[]}. Finally, the system matches the data from both observations and informs the user of the relevant shoe prices using \texttt{Finish[]}.

We adapted the original prompts to our task context and further refined them based on insights from preliminary experiments. The resulting prompts are available in Appendix~\ref{sec:react-prompts}. While the initial behavior is obtained through in-context learning, the following section introduces a novel pipeline for unsupervised fine-tuning of the \MakeLowercase{\model}s.

\section{Unsupervised Adaptation}
\label{sec:fine-tuning}

A key challenge in fine-tuning an agent for a custom environment is the lack of reasoning trajectory data specific to that setting. Although reinforcement learning could partially address this issue, it introduces substantial additional algorithmic complexity. Consequently, the main obstacle to straightforward adaptation lies in data scarcity. To overcome this limitation, we propose a novel approach, illustrated in Figure~\ref{fig:ft-diagram}. The figure shows a self-learning framework in which the model iteratively improves using its own generated reasoning trajectories. Our method consists of three main stages: generation, filtering, and fine-tuning.
We first leverage the in-context learning capabilities of the \MakeLowercase{\system} to automatically generate a training dataset containing \textit{correct} thought–action–observation trajectories. We assume that trajectories leading to correct final responses encapsulate valid reasoning patterns. After filtering for high-quality examples, we fine-tune the model on these automatically derived trajectories.

We employ an LLM-based judge to evaluate and select high-quality trajectories for fine-tuning automatically. To automate this filtering without human intervention, we employ Prometheus 2~\cite{kim2024prometheus2opensource}, an open-source LLM specifically aligned and trained to act as a fine-grained evaluator. It assesses the factual accuracy of the final responses, and we retain only those trajectories that receive a perfect score of 5 on its Likert scale. These threshold values were determined through preliminary experiments but can be regarded as tunable hyperparameters depending on the specific task and dataset.

Once the filtered dataset is obtained without any human supervision, we perform a supervised fine-tuning the model using LoRA~\cite{hu2021lora} to accommodate resource constraints. The best-performing checkpoint is selected during fine-tuning, taking Prometheus as the evaluation metric. We use the validation split for checkpointing. The model we obtain at this stage (first iteration) is ReAct FT.

Using Prometheus to evaluate the fine-tuned model allows us to incrementally expand the set of high-quality reasoning trajectories. The key idea behind our approach is that as the model improves through fine-tuning, it can, in turn, generate better trajectories than those produced in the initial \ICLSieben{} stage. We iteratively fine-tune the base model (Llama 8B) on the progressively enriched trajectory set until the process converges (i.e., when Prometheus detects no further performance gains). The final model obtained through this iterative process is referred to as Unsupervised Self-Improving ReAct (\UST{}). Figure~\ref{fig:ft-diagram} summarizes the complete unsupervised fine-tuning pipeline.

\section{Experimental setting}
\label{sec:exp-setting}

\paragraph{Dataset and hyperparameters} As noted in Section~\ref{sec:react-dialogue-system}, we evaluate our models on the SIMMC 2.1 dataset, which comprises 11,244 task-oriented dialogues totaling 117,236 utterances. Our experiments focus on dialogue response generation conditioned on the dialogue history and the available scene metadata. In our experiments, the models are provided with the two preceding dialogue turns as context.

The dataset is divided into two domains: fashion and furniture. The fashion domain contains a larger number of dialogues and exhibits substantially higher scene complexity. Specifically, fashion scenes include an average of 30 objects, compared to 8.9 objects in the furniture domain. Moreover, each fashion object is described by 11 attributes,\footnote{Fashion attributes: assetType, availableSizes, brand, color, customerReview, pattern, prefab\_path, price, size, sleeveLength, type.} whereas furniture objects have only 7.\footnote{Furniture attributes: brand, color, customerRating, materials, prefab\_path, price, type.} %
This difference in both object density and metadata richness significantly increases the load and context-window saturation for the models, making the fashion domain more challenging. We define this as ``scene complexity''.

Each dialogue in the dataset is paired with a corresponding scene characterized by a specific object composition and associated metadata. In addition, the dataset includes information for the utterances themselves, such as the intent.\footnote{The dialogue intent of the utterance is not used by the model. We use that information for analysis purposes.} We transformed the visual information from the rendered environments into textual metadata, in order to enable experimentation with unimodal models.

Based on preliminary analyses, we selected Llama 3.3 70B Instruct and Llama 3.1 8B Instruct~\cite{llama31} as the backbone models for the 70B and 8B versions of our ReAct \MakeLowercase{\system}s, respectively. For the ICL, each model is provided with five manually curated reasoning trajectories as few-shot examples. When fine-tuning, we employ LoRA~\cite{hu2021lora} with a rank of 32 to accommodate resource constraints.

\paragraph{Baselines} 
To evaluate the performance of our \MakeLowercase{\system}, we define two baseline approaches. The first, referred to as the \blind{} baseline, excludes all object-related information. The second, the \allcontext{} baseline, provides the model with the complete set of object metadata in the input. Both baselines utilize the Llama 3.3 70B Instruct model. The prompts are available in Appendix~\ref{sec:baselines-prompts}.

While evaluating the proposed \MakeLowercase{\system} against state-of-the-art models would be ideal, their training procedures render a direct comparison problematic. These models are frequently overfitted for specific datasets, a limitation that we intend to overcome with our proposal. Section~\ref{sec:generalizacion} explores these generalization constraints in depth, illustrating the inability of current SOTA methods to effectively generalize outside of their training distributions.

\paragraph{Evaluation} 
We opted not to use traditional automatic metrics (e.g., BLEU or BERTScore) for the final evaluation, as they fail to capture the factual nuances required for this task. Metrics designed for machine translation reward high lexical overlap; consequently, they may assign high scores to a generated response that is syntactically identical to the reference but hallucinates a crucial detail, such as stating a price of \$30 instead of \$300. At the same time, it could assign lower scores to fully correct answers that are shown in a different style. 
For the final evaluation, we adopted a blind manual annotation approach, in which 100 randomly selected instances per domain were evaluated without revealing the corresponding \MakeLowercase{\system}.

However, we still needed an automated heuristic for the filtering step described in Section~\ref{sec:fine-tuning}, so we evaluated the effectiveness of LLM judges for this task. Specifically, we adapted Prometheus 2~\cite{kim2024prometheus2opensource} to our evaluation setting. With the prompt available in Appendix~\ref{sec:prometheus-prompt}, we obtained a substantially higher correlation with manually annotated instances than traditional metrics such as BLEU4, which is used in the original dataset. However, we observed that the LLM-based judge tends to overestimate the performance of the baselines, leading to a higher rate of false positives.

Due to this limitation, the use of Prometheus is limited to filtering during the unsupervised self-improvement pipeline, where, additionally, we were able to check via manual evaluation that models did not overfit to Prometheus preferences. Moreover, it was used as a signal in some experiments, where the necessity for an automatic evaluation method was a must. However, in this cases, we took the necessary precautions to minimize Prometheus' biases by only comparing the same methods.

\section{Experimental Results}
\label{sec:main-results}

The \MakeLowercase{\system}s were evaluated using the blind manual assessment approach described in the previous section, judging the instances as correct or incorrect. The same evaluation instances were maintained across all \MakeLowercase{\system}s to ensure comparability. The results for each \MakeLowercase{\system} are shown in Table~\ref{tab:main-results}.

\begin{table}[!ht]
\centering
\resizebox{\columnwidth}{!}{%
    \begin{tabular}{lcc}
    \toprule
    \textbf{\system} & \textbf{\begin{tabular}[c]{@{}c@{}}Fashion\\ accuracy\end{tabular}} & \textbf{\begin{tabular}[c]{@{}c@{}}Furniture\\ accuracy\end{tabular}}\\
    \midrule
    \blind\ 70B & 30\% & 25\% \\
    \allcontext\ 70B & 50\% & 77\% \\
    \midrule
    \ICLAcht & 31\% & 42\% \\
    \ICLSieben & 60\% & 80\% \\ 
    \FTAcht & 65\% & 75\% \\
    \FTSieben & \textbf{68\%} & 76\% \\
    \midrule
    \UST & 67\% & \textbf{82\%} \\
    \bottomrule
    \end{tabular}
}
\caption{Results using manual evaluation for all \MakeLowercase{\system}s in both domains.}
\label{tab:main-results}
\end{table}

\begin{table*}[!tbp]
    \centering
    \resizebox{\textwidth}{!}{%
        \begin{tabular}{l|c|ccccc}
        \toprule
         & \textbf{Total} & \textbf{\blind} & \textbf{\allcontext} & \textbf{\begin{tabular}[c]{@{}c@{}}ReAct\\ ICL 70B\end{tabular}} & \textbf{\begin{tabular}[c]{@{}c@{}}ReAct\\ FT 70B\end{tabular}} & \textbf{\UST} \\
        \midrule
        \texttt{ASK:GET}               & 17 & 6\%  & 41\% & \textbf{76\%} & 65\% & 41\% \\
        \texttt{INFORM:DISAMBIGUATE}   & 16 & 0\%  & 50\% & 62\% & 69\% & \textbf{88\%} \\
        \texttt{INFORM:GET}            & 17 & 12\% & 59\% & 59\% & 65\% & \textbf{77\%} \\
        \texttt{INFORM:REFINE}         & 16 & 0\%  & 37\% & 62\% & 63\% & \textbf{75\%} \\
        \texttt{REQUEST:ADD\_TO\_CART} & 44 & 57\% & 59\% & 68\% & 84\% & \textbf{95\%} \\
        \texttt{REQUEST:COMPARE}       & 20 & 0\%  & 69\% & 65\% & 65\% & \textbf{70\%} \\
        \texttt{REQUEST:GET}           & 70 & 38\% & \textbf{80\%} & 77\% & 73\% & 67\% \\
        \bottomrule
        \end{tabular}
    }
    \caption{Correctly answered utterances filtered by the questions' intent.}
    \label{tab:results-UserActs}
\end{table*}

The results demonstrate that the ReAct 70B \MakeLowercase{\system}s outperform the baselines, highlighting the effectiveness of the ReAct approach. In general, the fine-tuned models surpass their respective in-context learning counterparts. Notably, \FTAcht{} achieves remarkable performance, significantly outperforming the \ICLSieben{} model in the fashion domain (65\% vs. 60\%), despite having significantly fewer parameters.

Interestingly, \UST{} also performs competitively against 70B models, surpassing \ICLSieben{} in both domains and achieving results comparable to \FTSieben{} in the fashion domain (-1\%) while substantially improving performance in the furniture domain (+8\%). This outcome is particularly notable given its smaller parameter size, which offers advantages such as reduced computational cost during inference, despite higher training requirements.

As expected, all \MakeLowercase{\system}s that use metadata perform better in the furniture domain than in the fashion domain, due to its simpler nature. This phenomenon also is present in the performance gap between the \allcontext{} baseline and \ICLSieben{}, which is smaller in the furniture domain than in the fashion domain. As discussed in Section~\ref{sec:objects}, this difference arises from the higher number of objects present in fashion scenes, which increases task complexity.

\section{Analysis}

\subsection{Intent Analysis}
\label{sec:error-analysis}

We know that user utterances vary in their reasoning demands (e.g., \textit{``compare the two tables''} vs. \textit{``add this table to the cart''}). To investigate \MakeLowercase{\system} performance, we leverage the intent annotations provided in the SIMMC 2.1 dataset to evaluate with respect to the user’s intent.

As described by \namecite{moon2020originalSIMMC}, each utterance (both user and system) is annotated with two levels of intent granularity. The higher-level intents, also referred to as \textit{dialogue acts}, fall into three categories: \texttt{ASK}, \texttt{INFORM}, and \texttt{REQUEST}. The \texttt{ASK} intent denotes information-seeking utterances, \texttt{INFORM} indicates those that provide information, and \texttt{REQUEST} corresponds to utterances requesting an action.

Each main intent is further subdivided into finer-grained sub-intents. Among them, the \texttt{GET} sub-intent is the most versatile, appearing under all three dialogue acts. It is used for requesting an item or retrieving its attributes. When combined with \texttt{REQUEST}, as in \texttt{REQUEST:GET}, it typically denotes generic item requests.\footnote{\texttt{REQUEST:GET}: \textit{Can you show me some shoes?}} In contrast, \texttt{ASK:GET} focuses on querying metadata information,\footnote{\texttt{ASK:GET}: \textit{What’s the price of that jacket?}} while \texttt{INFORM:GET} involves suggesting new items similar to others, often with additional constraints.\footnote{\texttt{INFORM:GET}: \textit{Do you have anything in a similar size to the grey shirt?}} The \texttt{INFORM:DISAMBIGUATE} sub-intent corresponds to utterances resolving prior ambiguity,\footnote{\texttt{INFORM:DISAMBIGUATE}: \textit{I meant the grey one.}} whereas \texttt{INFORM:REFINE} introduces further restrictions on an existing search.\footnote{\texttt{INFORM:REFINE}: \textit{Can we look for more expensive ones.}} Finally, \texttt{REQUEST:ADD\_TO\_CART} denotes adding an item to the cart,\footnote{\texttt{REQUEST:ADD\_TO\_CART}: \textit{Add the grey hat to my cart.}} and \texttt{REQUEST:COMPARE} indicates comparisons among items.\footnote{\texttt{REQUEST:COMPARE}: \textit{What sizes are available for each?}}

Table~\ref{tab:results-UserActs} presents \MakeLowercase{\system} accuracy across different user intents. As expected, the \blind{} baseline fails to handle questions requiring object-specific information (e.g., \texttt{REQUEST:COMPARE}, \texttt{INFORM:REFINE}). The \allcontext{} model shows a substantial improvement over this baseline, underscoring the importance of access to external information. It performs particularly well on queries involving multiple objects (e.g., \texttt{REQUEST:GET}, \texttt{REQUEST:COMPARE}). In contrast, \ICLSieben{} achieves notable gains on single-object reasoning tasks (e.g., \texttt{ASK:GET}) while maintaining accuracy on the remaining intents. \FTSieben{} improves performance across nearly all intent types and exhibits stronger alignment on simpler tasks relying primarily on contextual understanding (e.g., \texttt{REQUEST:ADD\_TO\_CART}). These gains are further amplified through the self-improvement loop, with \UST{} achieving the best overall results across most intents. Interestingly, \UST{} shows marked improvement in the \texttt{INFORM} category, demonstrating enhanced ability to leverage scene information to identify new relevant objects. However, its accuracy decreases in the \texttt{ASK:GET} intent, suggesting that future iterations of the pipeline should aim to preserve performance on simpler query-based reasoning tasks.

\subsection{Evolution of Self-Improving}

We assess the effectiveness of the self-improvement approach by tracking the increase in both the number of trajectories rated with a score of 5 and the average validation score assigned by Prometheus.
Table~\ref{tab:bootstrapEvolution} illustrates the progressive improvement of the model and the corresponding expansion of the trajectory set across iterations. In terms of the Prometheus score, the model improves from an initial value of 2.77 to 3.67 after five iterations. Likewise, the number of top-rated trajectories (score 5) increases substantially, from 5,492 in the fashion domain (and 2,342 in furniture) at the initial stage to 11,073 in fashion (and 4,408 in furniture) after the final iteration. Note that the increasing ratio is similar in the two domains.

The analysis reveals that the first iteration yields the largest improvement, increasing the number of filtered instances by 55\% in the fashion domain and 43\% in furniture. Subsequent iterations result in smaller gains, indicating that the initial fine-tuning step plays a crucial role in adapting the \MakeLowercase{\model}s. This trend is consistent with the performance of \FTAcht{}, as shown in Table~\ref{tab:main-results}, where a single fine-tuning iteration leads to substantial performance gains.
We stopped training after 5 iterations, as it showed worse performance in the 6\textsuperscript{th} fine-tuning step.

The Prometheus score analysis was conducted exclusively on the validation partition of the fashion domain; thus, we acknowledge that the findings may be limited in scope. Nevertheless, it is worth noting that fashion represents the most challenging domain in our setup, and the observed trends are expected to generalize to the furniture domain as well.

\begin{table*}[htbp]
\centering
\begin{tabular}{l|ccccccc}
\toprule
\textbf{Iteration} & \textbf{Initial} & \textbf{1} & \textbf{2} & \textbf{3} & \textbf{4} & \textbf{5} & \textbf{6} \\ \midrule
Fashion utterances & 5492 & 8542 & 9653 & 10289 & 10789 & \textbf{11073} & - \\
Furniture utterances & 2342 & 3368 & 3893 & 4117 & 4206 & \textbf{4408} & - \\
Prometheus score (val) & 2.77 & 3.51 & 3.56 & 3.59 & 3.63 & \textbf{3.67} & 3.63 \\ \bottomrule

\end{tabular}%
\caption{Training set evolution throughout the self-improvement process. This table shows the number of high-quality training utterances (Prometheus scoring 5/5) used for each iteration, along with the mean Prometheus score on the validation set. Selected iteration's results in bold.}
\label{tab:bootstrapEvolution}
\end{table*}

\subsection{Effect of Scene Complexity}
\label{sec:objects}

We analyze how scene complexity, measured by the number of included objects, influences model performance. We investigate whether (1) the higher performance observed in the furniture domain arises from its smaller number of objects, and (2) whether the \allcontext{} baseline may struggle as the number of objects increases substantially.

We conducted an additional analysis by augmenting the validation set with synthetic (fake) objects in the scene metadata. Multiple dataset variants were created, each containing a different number of total objects per scene, ranging from a minimum of 20 up to 150. For dialogues with fewer objects than the target count, we inserted fake objects into the metadata. These synthetic entries were generated using attribute values not originally present in the scene, such as novel colors, sizes, and clothing categories, to ensure clear differentiation from real objects.

\begin{figure}[!ht]
\begin{center}
\includegraphics[width=\columnwidth]{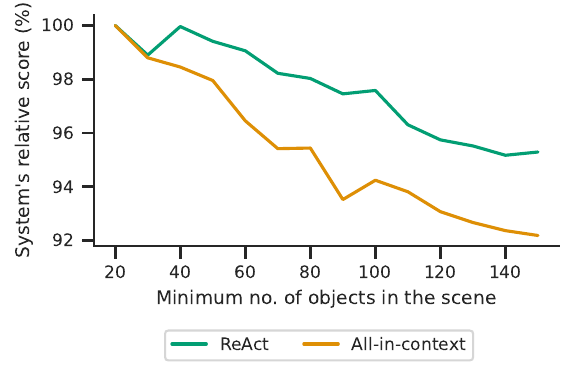}
\caption{Prometheus average scores for the different \MakeLowercase{\system}s as the number of objects increases in the fashion domain.}
\label{fig:fake-objects}
\end{center}
\end{figure}

The results of this analysis are presented in Figure~\ref{fig:fake-objects}. It shows the performance gap between \ICLSieben{} and the \allcontext{} baseline. Notably, as the number of objects increases, the performance of the \allcontext{} baseline degrades substantially. In contrast, \ICLSieben{} exhibits greater robustness under complex conditions. These findings reinforce the effectiveness of the reasoning and tool-use framework in scenarios that require managing wide context windows and highly specific situational information.

\subsection{Filtering noisy triplets}
\label{sec:filtering-triplets}
Experimental results indicate that Prometheus effectively identifies high-quality reasoning trajectories, enabling smaller fine-tuned models to outperform larger ICL-based models. It is important to note, however, that the LLM evaluates each trajectory holistically, rather than assessing the quality of individual \textit{thought–action–observation} triplets within the reasoning process. This limitation may influence the outcome of fine-tuning, as the model could inadvertently learn to produce longer or inefficient trajectories that ultimately lead to incorrect responses.

We define a simple heuristic that filters out incorrect triplets. The filter discards malformed actions, such as incomplete operations or invalid search queries, and identifies multiple consecutive uses of the \texttt{Look[]} tool as inefficient, since they consistently yield identical results.

\begin{table}[!ht]
\centering
\begin{tabular}{lcc}
\toprule
\textbf{\system} & \textbf{\begin{tabular}[c]{@{}c@{}}Filtered\\ trajectories\end{tabular}} & \textbf{\begin{tabular}[c]{@{}c@{}}W/o\\ filtering\end{tabular}} \\ \midrule
\FTSieben & \textbf{2.8951}\scriptsize{(-0.67)} & 3.5608 \\
\FTAcht & \textbf{2.7906}\scriptsize{(-0.53)} & 3.3172 \\
\UST & \textbf{2.4974}\scriptsize{(-0.29)} & 2.7849 \\
\bottomrule
\end{tabular}
\caption{Mean number of triplets generated by each \MakeLowercase{\system} based on whether triplet filtering was applied. Lower is better.}
\label{tab:mean-triplets}
\end{table}

Table~\ref{tab:mean-triplets} shows that the number of generated triplets decreases substantially when the model is fine-tuned on cleaned trajectories. This reduction is particularly pronounced for models that initially produce a large number of triplets.
Interestingly, the non-filtered \UST{} model generates relatively few triplets, suggesting that the iterative self-learning framework implicitly learns to discard trajectories containing noisy or uninformative triplets.

\begin{table}[!ht]
\centering
\begin{tabular}{lcc}
\toprule
\textbf{\system} & \textbf{\begin{tabular}[c]{@{}c@{}}Filtered\\ trajectories\end{tabular}} & \textbf{\begin{tabular}[c]{@{}c@{}}W/o\\ filtering\end{tabular}} \\ \midrule
\FTSieben & 72\% & \textbf{74\%} \\
\FTAcht & \textbf{70\%} & 62.5\% \\
\UST & \textbf{74.5\%} & 73.5\% \\
\bottomrule
\end{tabular}
\caption{Results of fine-tuned \MakeLowercase{\system}s based on whether triplet filtering was applied.}
\label{tab:results-based-filter}
\end{table}

The results shown in Table~\ref{tab:results-based-filter} highlight the importance of using clean trajectories. Most models trained on the filtered dataset exhibit considerable improvements. This is particularly pronounced in the \FTAcht{} model, improving the model up to 7 points.

\subsection{Generalization of Fine-tuned ReAct}
\label{sec:generalizacion}

We evaluate the cross-domain generalization and transferability of our fine-tuning pipeline. This analysis is particularly relevant, as fine-tuned models can easily memorize object properties within a domain, leading to overfitting~\cite{hemanthage2023learnmetadata}. To assess this, we train the models on one domain and evaluate them on both. We further compare our fine-tuned model against a BART state-of-the-art architecture in this task~\cite{lee-etal-2022-learning,lewis-etal-2020-bart}, and use it as a baseline to check for possible memorization differences between the architectures.

\begin{table}[!ht]
\centering
\resizebox{\columnwidth}{!}{%
\begin{tabular}{llcc}
\toprule
\textbf{System} & \textbf{\begin{tabular}[c]{@{}l@{}}Trained\\ on\end{tabular}} & \textbf{\begin{tabular}[c]{@{}c@{}}Fashion\\ accuracy\end{tabular}} & \textbf{\begin{tabular}[c]{@{}c@{}}Furniture\\ accuracy\end{tabular}} \\ \midrule
\multirow{2}{*}{\External} & Fashion & \textbf{19\%} & 10\% \\
                           & Furniture & 7\% & \textbf{15\%} \\ \midrule
\multirow{2}{*}{\FTAcht} & Fashion & 21\% & \textbf{22\%} \\
                         & Furniture & 20\% & \textbf{25\%} \\ \bottomrule
\end{tabular}%
}
\caption{Percentage of instances achieving a Prometheus evaluation score of 3 or higher. The models were trained only on a specific domain and evaluated in both.}
\label{tab:generalization}
\end{table}

The results of these experiments are presented in Table~\ref{tab:generalization}. Our fine-tuning approach demonstrates superior cross-domain generalization, as the performance drop observed during cross-domain evaluation is substantially smaller than that of the BART baseline. Interestingly, our model trained on the fashion domain achieves higher scores when evaluated on furniture, consistent with prior observations that domain complexity influences performance. In contrast, the BART models exhibit a pronounced decline in performance when tested on out-of-domain instances.

\section{Conclusions}

In this work, we demonstrated the successful adaptation of the ReAct framework to task-oriented dialogue systems, mitigating hallucination issues by grounding the model in its environment through dynamic information retrieval. Furthermore, we showed that an effective unsupervised fine-tuning pipeline can be developed to leverage reasoning trajectories without relying on human feedback.

Our experiments on the SIMMC dataset demonstrate that the proposed approach not only outperforms baseline LLMs that rely solely on in-context information, but also exhibits greater robustness in complex scenarios characterized by high object density. Importantly, our self-improvement pipeline enables a Llama 3 8B model to match or even surpass the performance of a Llama 3 70B in-context system. %
Moreover, our analysis shows that filtering inefficient reasoning steps improves both inference efficiency and overall performance, while fine-tuning on reasoning trajectories fosters strong cross-domain generalization.

\section{Limitations}

While our approach demonstrates the effective adaptation of the ReAct framework to dialogue systems, several limitations remain, highlighting opportunities for future research.

First, our experiments are conducted exclusively on the SIMMC 2.1 dataset. Although this dataset provides a rich, multimodal environment with multiple domains, extending the evaluation to a broader range of task-oriented dialogue datasets would enable a more comprehensive assessment of the  proposed approach's robustness.

Second, despite the multimodal nature of the SIMMC 2.1 dataset, our method utilizes only textual and structured metadata, omitting the direct use of visual input. Integrating visual signals directly into the ReAct framework remains an important direction for future research.

Finally, our experimental results rely primarily on manual evaluation, which limits the assessment to a few hundred examples. This highlights the critical need for developing reliable automatic evaluation methods, which would enable a more scalable analysis of system performance across larger sets of examples.

\section*{Acknowledgments}

This work was developed under the grant 101135724 (LUMINOUS) of the EU Horizon Europe Framework.

We wish to thank many of our colleagues in the research group for their support, including technical and administrative assistance, throughout this project. A special thank you goes to Oier Ijurco for his valuable insights into the dataset and his participation in shaping ideas behind this work.

\bibliographystyle{fullname}
\bibliography{EjemploARTsepln}
\label{sec:reference}

\appendix
\section{Prompts}

\newtcolorbox[blend into=figures]{promptbox}[2][]{%
colback=gray!1!white, breakable,pad at break*=1mm, fonttitle=\bfseries, title=#2.,#1}

\subsection{Baselines prompts}
\label{sec:baselines-prompts}

\begin{promptbox}{\blind{} baseline prompt}
You are a shop assistant that must answer the user's question. Give your answer in a single sentence.\newline\newline
Conversation: \{Previous turns and current user question\}

System:
\end{promptbox}

\begin{promptbox}{\allcontext{} baseline prompt}
You are a shop assistant that must answer the user's question. Give your answer in a single sentence.
\newline\newline
Available items' visual features:

\{Visual metadata of all objects in the scene\}
\newline\newline
Available items' hidden features:

\{Hidden metadata of all objects in the scene\}
\newline\newline
Conversation: \{Previous turns and current user question\}

System: 
\end{promptbox}

\subsection{ReAct \MakeLowercase{\system}s prompts}
\label{sec:react-prompts}

\begin{promptbox}{\ICLSieben{} fashion prompt}
You are a seller in a fashion shop. A client comes to you and you must answer their questions. Solve the questions answering task with interleaving Thought, Action, Observation steps. Thought can reason about the current situation, and Action can be three types:

(1) Search[query], which searches the existing catalogue. You may search for customerReview, availableSizes, brand, price, size or prefab\_path. You MUST NOT use other parameters or the query will error. You may use the following operators: ==, <=, <, >=, >, !=, in, and, or.

(2) Look[], which returns the items that are visible from your perspective. The position, assetType, color, pattern, sleeveLength, type and prefab\_path will become apparent to you.

(3) Finish[answer], which returns the answer to the client and finishes the task.

Here are some examples (DO NOT use information from example conversations in other conversations; you must ask again for the information):

\{Few-shot examples\}

\end{promptbox}

\begin{promptbox}{\ICLSieben{} furniture prompt}
You are a seller in a furniture shop. A client comes to you and you must answer their questions. Solve the questions answering task with interleaving Thought, Action, Observation steps. Thought can reason about the current situation, and Action can be three types:

(1) Search[query], which searches the existing catalogue. You may search for prefab\_path, brand, price, customerRating or materials. You MUST NOT use other parameters or the query will error. You may use the following operators: ==, <=, <, >=, >, !=, and, or.

(2) Look[], which returns the items that are visible from your perspective. The position, corner\_pos, type, color and prefab\_path will become apparent to you.

(3) Finish[answer], which returns the answer to the client and finishes the task.

Here are some examples (DO NOT use information from example conversations in other conversations; you must ask again for the information):

\{Few-shot examples\}
\end{promptbox}

\subsection{Prometheus Evaluation prompt}
\label{sec:prometheus-prompt}

\begin{promptbox}{Prometheus Evaluation prompt}
\#\#\#Task Description:

An instruction (might include an Input inside it), a response to evaluate, a reference answer that gets a score of 5, and a score rubric representing a evaluation criteria are given.

1. Write a detailed feedback that assess the quality of the response strictly based on the given score rubric, not evaluating in general.

2. After writing a feedback, write a score that is an integer between 1 and 5. You should refer to the score rubric.

3. The output format should look as follows: "(write a feedback for criteria) [RESULT] (an integer number between 1 and 5)"

4. Please do not generate any other opening, closing, and explanations.

\#\#\#The instruction to evaluate:

\#\#\# Relevant objects:

\{objects\}

\#\#\# Conversation:

\{question\}

\#\#\#Response to evaluate:

\{final\_answer\}

\#\#\#Reference Answer (Score 5):

\{correct\_answer\}

\#\#\#Score Rubrics:

[Does the model correctly represent the relevant object\{plural\} factually?]

Score 1: The model answers with incorrect information or information about objects that are not relevant.

Score 2:

Score 3:

Score 4:

Score 5: The model answers the question and provides accurate information about the relevant object(s).

\#\#\#Feedback: 
\end{promptbox}

\end{document}